\documentclass[letterpaper]{article} 
\usepackage{aaai23}  
\usepackage{times}  
\usepackage{helvet}  
\usepackage{courier}  
\usepackage[hyphens]{url}  
\usepackage{graphicx} 
\usepackage{multirow}
\usepackage{natbib}  
\usepackage{caption} 
\usepackage{algorithm}
\usepackage{algorithmic}
\usepackage{xcolor}
\newcommand{\answerYes}[1]{\textcolor{blue}{#1}} 
 
\newcommand{\answerNA}[1]{\textcolor{gray}{#1}} 
 
\usepackage{newfloat}
\usepackage{listings}
\DeclareCaptionStyle{ruled}{labelfont=normalfont,labelsep=colon,strut=off} 
\floatstyle{ruled}
\newfloat{listing}{tb}{lst}{}
\floatname{listing}{Listing}
\title{Stance Detection with Collaborative Role-Infused LLM-Based Agents}
\author{
    Xiaochong Lan,
    Chen Gao$^*$,
    Depeng Jin,
    Yong Li$^*$
}
\affiliations{Department of Electronic Engineering, BNRist, Tsinghua University, China

    lanxc22@mails.tsinghua.edu.cn, chgao96@gmail.com, jindp@tsinghua.edu.cn, liyong07@tsinghua.edu.cn}

\usepackage{bibentry}

\begin{document}

\maketitle
\begin{abstract}
\let\thefootnote\relax\footnotetext{*Corresponding Author}
Stance detection automatically detects an author's position on a particular topic within a text, vital for content analysis in web and social media research. 
With the development of LLMs, researchers have begun to explore their potential for stance detection. Despite their promising capabilities, LLMs encounter challenges when directly applied to stance detection. First, stance detection demands multi-aspect knowledge to fully understand elements in the text. Second, stance detection requires advanced reasoning to infer viewpoints, as stances are often implicitly embedded rather than explicitly stated in the text. To address these challenges, we design a three-stage framework COLA (short for \textbf{C}ollaborative r\textbf{O}le-infused \textbf{L}LM-based \textbf{A}gents) in which LLMs are designated distinct roles, creating a collaborative system. The framework consists of three stages. First, in the multidimensional text analysis stage, we configure the LLMs to act as a linguistic expert, a domain specialist, and a social media veteran to get a multifaceted analysis of texts, thus overcoming the first challenge. Next, in the reasoning-enhanced debating stage, for each potential stance, we designate a specific LLM-based agent to advocate for it, guiding the LLM to detect logical connections between text features and stance, addressing the second challenge. Finally, in the stance conclusion stage, a final decision maker agent consolidates prior insights to determine the stance. COLA avoids the need for extra annotated data and model training, making it highly user-friendly. What's more, COLA achieves state-of-the-art performance across multiple widely-used datasets. Ablation studies validate the effectiveness of each module in our approach. Further experiments have demonstrated the explainability and the versatility of our approach. In summary, our approach excels in usability, accuracy, effectiveness, explainability and versatility, showcasing its significant value.
\end{abstract}

\section{Introduction}
Stance detection is commonly defined as automatically detecting the stance (as \textit{Favor}, \textit{Against}, or \textit{Neutral}) of the author towards a target~\cite{mohammad2016semeval}. Over the years, numerous methodologies have been proposed for stance detection~\cite{kuccuk2020stance,aldayel2021stance}. However, a persistent challenge lies in the need to train models specifically for the targets of interest. Even with advancements in cross-target stance detection~\cite{liang2021target} and zero-shot stance detection~\cite{allaway2020zero,liang2022zero}, training on annotated corpora is always required. However, acquiring large-scale labeled datasets is not trivial, which curtails the model's usability.

Recently, large language models (LLMs) have demonstrated remarkable capabilities across various applications~\cite{brown2020language,park2023generative}. The inherent semantic understanding of these large models presents an exciting opportunity for stance detection. Most LLMs can be easily interacted with by users through zero-shot prompting, which significantly enhances their usability. Thus, with their strength and usability, large language models offer new possibilities for stance detection.

\begin{figure}[ht]
    \centering
    \includegraphics[width=0.95\linewidth]{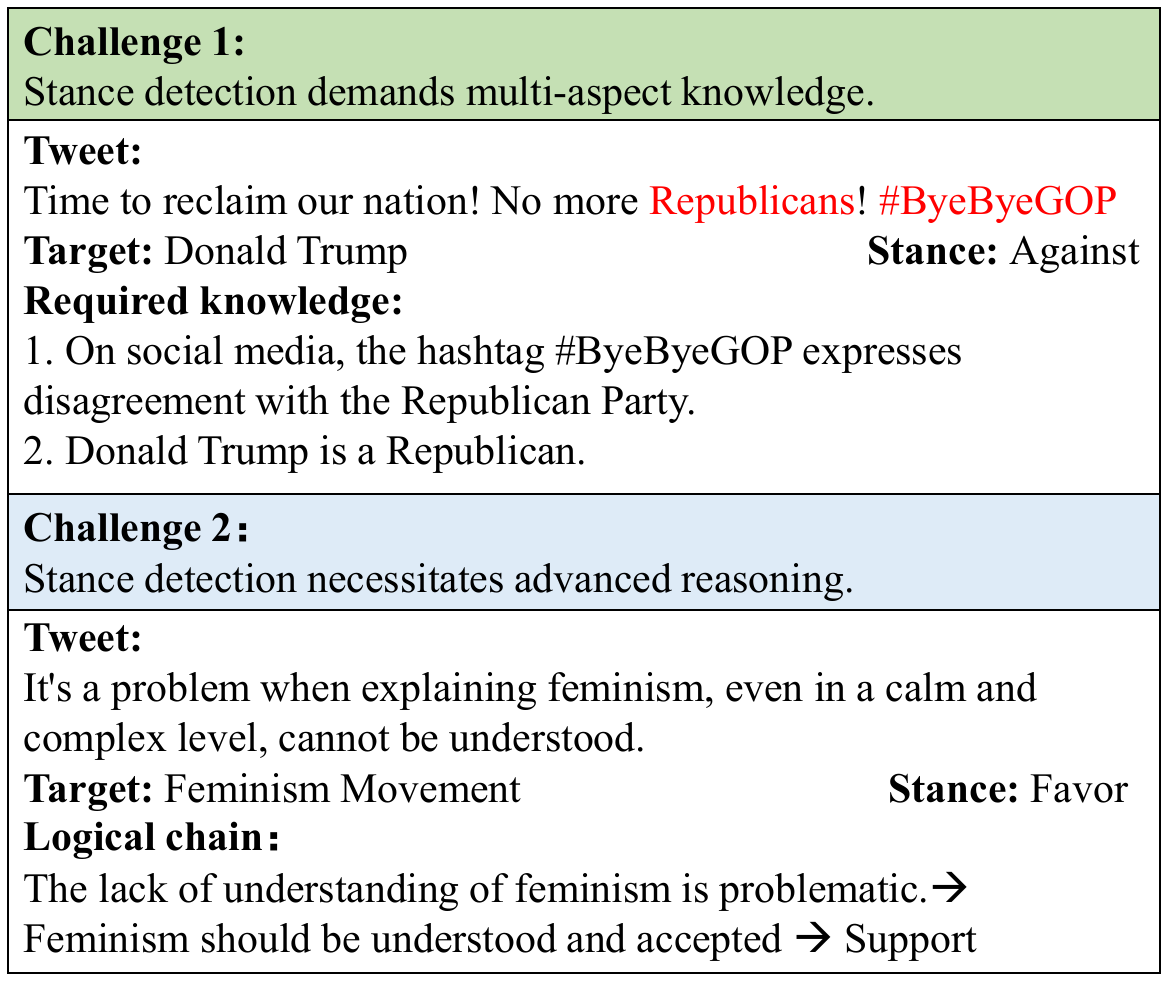}
    \caption{Illustration of the challenges of stance detection.}
    \label{fig:challenge}
\end{figure}

Researchers have discerned the transformative potential LLMs bring to stance detection. Some works have proposed simple methods using LLMs for stance detection~\cite{zhang2022would,zhang2023investigating}. Yet, while these works report satisfactory results on specific subsets of certain datasets, our rigorous replications indicate that these methods frequently fail to match the performance of state-of-the-art non-LLM baselines. This can be attributed to two inherent challenges of stance detection, which can be listed as follows. 
\begin{itemize}
\item{\textbf{First, stance detection demands multi-aspect knowledge.} As shown in Figure~\ref{fig:challenge}, sentences may contain elements like domain-specific terms, cultural references, social media linguistic styles, and more. These are not immediately comprehensible to large language models and require specialized parsing to be truly understood.}

\item{\textbf{Second, stance detection necessitates advanced reasoning.} Often, authors don't state their stances directly but inadvertently reveal them in various ways, such as through their attitudes towards related topics or events, as shown in Figure~\ref{fig:challenge}. Stance detection requires reasoning from various textual features to arrive at the correct stance.}
\end{itemize}

To address these challenges, we introduce our three-stage framework named COLA (short for \textbf{C}ollaborative r\textbf{O}le-infused \textbf{L}LM-based \textbf{A}gents). Specifically, we design a stance detection system consisting of role-infused LLM-based agents, with each role bearing distinct responsibilities and significance. To counter the first challenge, we design a multidimensional text analysis stage. In this stage, LLMs are designated with three roles, named as linguistic expert, domain specialist, and social media veteran, to analyze text from various perspectives, covering syntax, textual elements, and platform-specific expressions, ultimately revealing stance indicators. Addressing the second challenge, we propose a reasoning-enhanced debating stage. In this stage, advocates for each stance category draw evidence from previous analyses, presenting arguments that compel LLMs to uncover the underlying logic linking textual features and stances. Lastly, a stance conclusion stage determines the text's stance, drawing insights both from the original text and the debates.

Our approach does not necessitate annotated data nor additional model training, hence ensuring high \textbf{usability}. Extensive experiments validate our method's superior performance over existing baselines, affirming its \textbf{accuracy}\footnote{In this work, unless explicitly stated otherwise, we use \textit{accuracy} to express the overall strong performance of the model on classification tasks, rather than solely referring to the accuracy metric.}. Ablation studies demonstrate the \textbf{effectiveness} of each module. Case studies and quantitative experiments show that our approach can generate reasonable explanations for its output, demonstrating our approach's \textbf{explainability}. The powerful performance of our proposed framework in a series of text classification tasks underscores its \textbf{versatility}. Our approach stands out for its usability, accuracy, effectiveness, explainability, and versatility, all of which highlight its value.

Our main contributions are summarized as follows:
\begin{itemize}
\item{To the best of our knowledge, we are the first to employ multiple LLM agents for stance detection.}
\item{We introduce an approach based on collaborative role-infused LLM-empowered agents, which achieves a remarkable 19.2\% absolute improvement over the best non-LLM zero-shot stance detection baseline on the SEM16 dataset. Additionally, it offers high usability and explainability.}
\item{Our proposed three-stage framework—analyst, debater, and summarizer—offers significant potential for a range of text classification tasks, providing a powerful tool for text analysis on web and social media.}
\end{itemize}

The subsequent sections are organized as follows. In Section~\ref{sec::related_work}, we review related works. In the Section~\ref{sec::methods}, we describe our three-stage framework in detail. Then, in Section~\ref{sec::experiments} and \ref{sec::results}, we present our experiments, providing robust empirical evidence that demonstrates the superiority of our method from multiple perspectives. Lastly, in Section~\ref{sec::conclusion}, we conclude our work and highlight potential areas for future improvement.

\section{Related Work}\label{sec::related_work}
This section is structured as follows: First, we provide a detailed overview of advancements in stance detection. Next, we introduce recent progress in large language models. Lastly, we focus on reviewing a subset of works closely related to ours, specifically multi LLM-based agents systems.

\textbf{Stance detection.}
Stance detection aims to discern the stance of the author towards a particular target from textual content. Typically, stances are categorized into favor, against, neutral. A plethora of algorithms for stance detection have been proposed by researchers, encompassing both feature-based methods~\cite{bar2017stance,lozhnikov2020stance} and deep learning techniques~\cite{wei2018target,liu2021enhancing}. These methodologies have enabled in-depth analysis of content on the internet and social media platforms. For example, Jang et al.~(\citeyear{jang2018explaining}) develop a method to find controversies on social media by generating stance-aware summaries of tweets. Grcar et al.~(\citeyear{grvcar2017stance}) examine the Twitter stance before the Brexit referendum, revealing the pro-Brexit camp's higher influence.

Conventionally, stance detection necessitates training on datasets annotated for specific targets. Such datasets are not trivially obtainable, thereby constraining the usability of many methods. Recognizing this limitation, researchers have ventured into cross-target stance detection, aiming to train classifiers that can adapt to unfamiliar but related targets after being trained on a known target~\cite{xu2018cross,wei2019modeling,liang2021target}. Recently, there has been an emergence of zero-shot stance detection approaches that automatically detects the stance on unseen tasks~\cite{allaway2020zero,liang2022zero}. However, all these methods require training on annotated datasets. Unlike these methods, our approach uses pre-trained LLM, removing the need for additional annotated data. Through prompt engineering, we refine these models without extra training, offering a solution with high usability.

\textbf{Large language models.}
Large language models (LLMs) represent one of the most significant advancements of artificial intelligence in recent years. Since the release of ChatGPT\footnote{chat.openai.com} at the end of 2022, LLMs have witnessed a meteoric rise in attention, predominantly driven by their outstanding performance. A myriad of LLMs, such as GPT-4~\cite{openai2023gpt}, Llama 2~\cite{touvron2023llama}, ChatGLM~\cite{zeng2022glm}, and others, have been introduced at a rapid pace. In conventional NLP tasks, the zero-shot capabilities of these LLMs often rival or even surpass meticulously crafted, domain-specific models~\cite{wei2021finetuned}. The emergence of robust capabilities, such as planning and reasoning within LLMs, has further enabled their adoption across diverse applications. Some endeavors integrate LLMs with existing tools~\cite{qin2023toolllm,schick2023toolformer}, others explore the potential of LLMs to create new tools~\cite{cai2023large}, and there is a growing trend towards leveraging LLMs for dynamic decision-making, planning, and embodied intelligence~\cite{shinn2023reflexion,xiang2023language}.

Inherently, the vast knowledge and potent semantic understanding of LLMs offer immense potential in tackling stance detection tasks. Several research initiatives have indeed explored the application of LLMs in stance detection~\cite{zhang2022would,ziems2023can,zhang2023investigating}. However, these existing methods often adopt relatively straightforward approaches, neglecting the intrinsic challenges specific to stance detection. As a result, our rigorous replication efforts have frequently found their performance to be subpar in comparison to annotated data dependent baselines. In contrast, our method is specifically tailored to cater to the expert knowledge and intricate reasoning often required for stance detection, consequently achieving commendable results.

\textbf{Multi LLM-based agents system.}
Systems comprised of multiple LLM-based agents have demonstrated complex and powerful capabilities not inherent to individual LLM. Leveraging the human-like capacities of LLM, systems formed from multiple LLM-based agents have been applied in both online and offline societal simulations, showcasing credibility at the individual level and emergent social behaviors~\cite{li2023large,gao2023large}. For instance, Part et al.~(\citeyear{park2023generative}) construct an AI town with 25 agents, witnessing phenomena such as mayoral elections and party organization. Gao et al.~(\citeyear{gao2023s}) conduct simulations of online social networks with thousands of LLM-based agents, observing group emotional responses and opinion shifts that mirrored real-world trends. What's more, some studies have employed collaborative efforts between LLMs with distinct roles to accomplish tasks. In METAGPT~\cite{hong2023metagpt}, LLM-based agents with different roles collaboratively develop computer software, while DERA~\cite{nair2023dera} uses discussions among various agents to refine medical summary dialogues and care plan generation. Additionally, several efforts have utilized debates between large language model agents to enhance model performance. For example, ChatEval~\cite{chan2023chateval} improves text evaluation capabilities through multi-agent debates. Du et al.(\citeyear{du2023improving}) amplify the factuality and reasoning capacities of large language models by facilitating debates among them. 

To the best of our knowledge, our work is the pioneering effort in employing multi LLM-based agents for the task of stance detection.

\begin{figure*}[]
    \centering
    \includegraphics[width=0.95\textwidth]{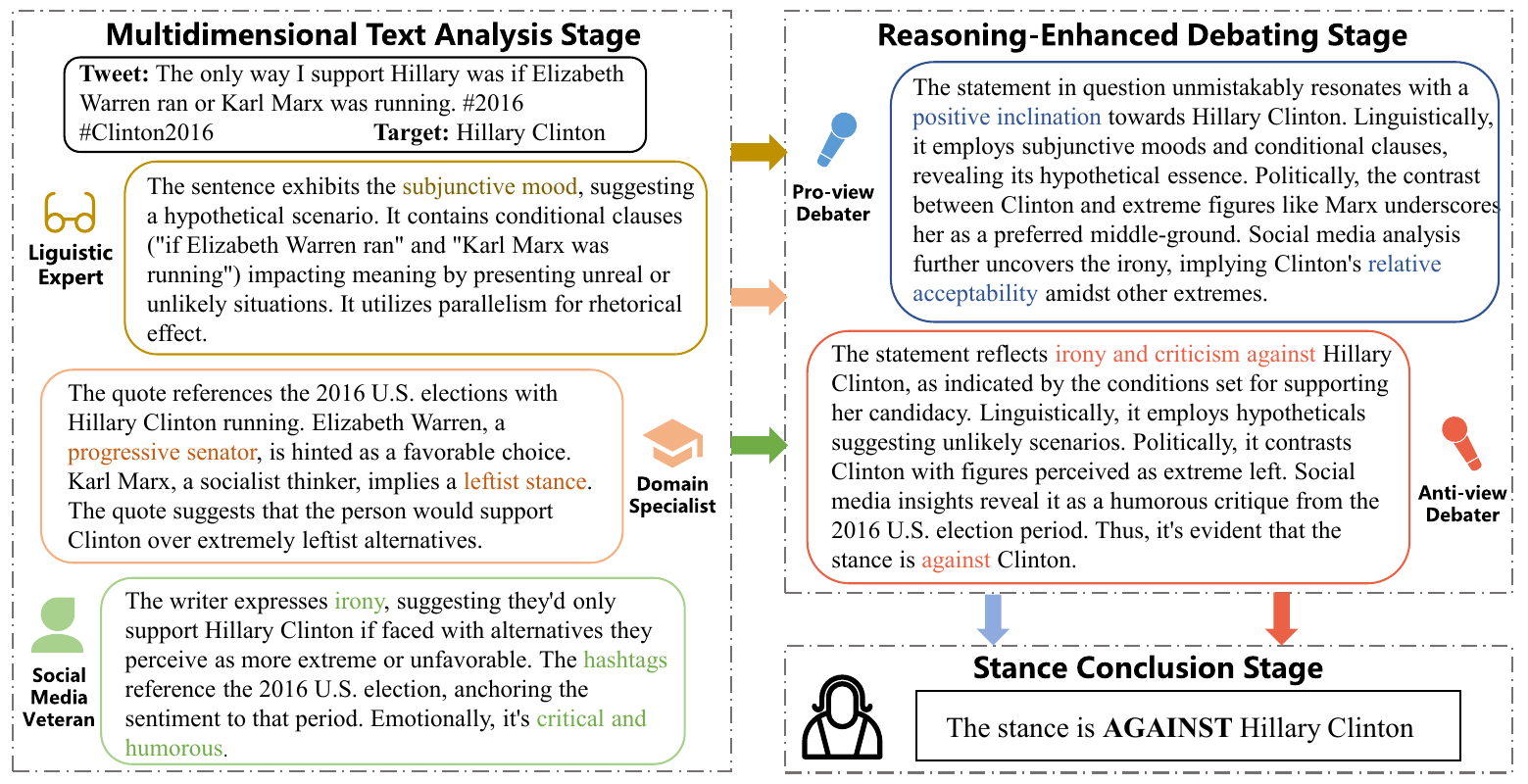}
    \caption{Architecture of our proposed COLA.}
    \label{fig:framework}
\end{figure*}

\section{Methods}\label{sec::methods}
\subsection{Task Description and Model Overview}

In stance detection, the objective is to decide the stance of a given opinionated document with respect to a specified target. Let us define a dataset $D=\{(x_i=(d_i, t_i), y_i)\}^{n}_{i=1}$ consisting of $n$ instances. For each instance, $x_i$ represents a tuple comprising a document $d_i$ and a corresponding target $t_i$. The task is to detect the stance $y_i$, which can be one of the following categories: favor, against, or neutral.

As illustrated in Figure~\ref{fig:framework}, our approach consists of three stages: multidimensional text analysis stage, reasoning-enhanced debating stage, and stance conclusion stage. In the multidimensional text analysis stage, the linguisic expert, the domain specialist and the social media veteran analyze the text from web or social media from various perspectives, providing a holistic understanding. In the reasoning-enhanced debating stage, for each possible stance, a debater defends it, seeking possible logical chains between text features and the stance. Finally, in the stance conclusion stage, a final judge determines the stance based on the statements made by all debaters. Next, we will introduce the components of our approach in detail.
\subsection{Multidimensional Text Analysis Stage}
\subsubsection{Challenge:} Stance detection necessitates a profound grasp of multi-aspect knowledge. Sentences on social media that convey the authors' stances may be influenced by various linguistic phenomena, such as grammatical structures, tenses, and moods. There is also often an abundance of domain-specific terminologies, including references to characters, political parties, and events, and their relationships with the target. Additionally, unique language features of social media, such as hashtags, come into play. Although large language models have assimilated vast knowledge from their training data, their direct application for stance detection often fails to adequately harness this knowledge, leading to suboptimal results, a fact corroborated by our subsequent experiments.

\subsubsection{Approach:} To address this challenge and leverage the rich knowledge encoded within large language models, we designed a multidimensional text analysis stage. During this stage, we introduce three distinct LLM-based agents to parse the text from different perspectives, ensuring a comprehensive understanding of potential elements influencing the author's stance. These agents are the Linguistic Expert, Domain Specialist, and Social Media Veteran. We ask the LLM to behave as their designated roles through prompting.
Specifically, the inputs and outputs of the role-infused agents in this stage are as follows.

\textbf{Input:} A text with a stance. 

\textbf{Output:} The individual analyses of the text by the linguistic expert, the domain specialist, and the social media veteran.

The detailed configurations of agents are as follows.

\textbf{Linguistic Expert.} This Agent is tasked with dissecting the text from a linguistic standpoint, exploring factors including but not limited to:
\begin{itemize}
\item{\textit{Grammatical structure.} The arrangement and relationship of words in a sentence, which determines how different elements combine to produce specific meanings.}
\item \textit{Tense and inflection.} Tense identifies when an action occurs, influencing the stance's immediacy or distance. Inflection adjusts word forms, providing clues about the sentence's grammatical and relational context.
\item \textit{Rhetorical devices.} These are techniques used to enhance the expressiveness of language. By emphasizing, contrasting, or evoking emotions, they shape the tone and attitude of a statement. 
\item{\textit{Lexical choices.} The selection of particular words or phrases in writing, which can reveal deeper nuances, biases, or viewpoints about a topic.}
\end{itemize}
The specific prompt is as follows,

\begin{quote}
    \setlength{\leftskip}{0.3cm}
    \setlength{\rightskip}{0.3cm}
    \textit{You are a linguist. Accurately and concisely explain the linguistic elements in the sentence and how these elements affect meaning, including grammatical structure, tense and inflection, virtual speech, rhetorical devices, lexical choices and so on. Do nothing else. \{tweet\}}
\end{quote}

\textbf{Domain Specialist.} This agent focuses on domain-relevant knowledge, exploring facets such as:
\begin{itemize}
\item{\textit{Characters.} Key individuals or entities in a text.}
\item{\textit{Events.} Significant occurrences within a text. How they're portrayed can hint at the author's stance on certain issues or topics.}
\item{\textit{Organizations.} Established groups mentioned. Their depiction can showcase the author's feelings towards certain societal structures or institutions.}
\item{\textit{Parties.} Political groups with distinct ideologies. A text's treatment of these can provide insights into the author's political leanings or criticisms.}
\item{\textit{Religions.} Specific faiths or spiritual beliefs. How they are referenced might shed light on the author's personal beliefs or societal observations.}
\end{itemize}
The specific prompt is as follows,
\begin{quote}
    \setlength{\leftskip}{0.3cm}
    \setlength{\rightskip}{0.3cm}
\textit{You are a \{role\}. Accurately and concisely explain the key elements contained in the quote, such as characters, events, parties, religions, etc. Also explain their relationship with \{target\} (if exist). Do nothing else. \{tweet\}}
\end{quote}

\textbf{Social Media Veteran.} This agent delves into the nuances of social media expression, focusing on aspects like:
\begin{itemize}
\item{\textit{Hashtags.} Specific labels used on social media platforms, assisting in categorizing posts or emphasizing specific themes, making content easily discoverable.}
\item{\textit{Internet slangs and colloquialisms.} These refer to informal terms and expressions often used in online communities. Their usage can introduce nuances, cultural contexts, or specific attitudes, making them significant indicators of the underlying stance in a statement.}
\item{\textit{Emotional tone.} This captures the sentiment inherent in a piece of writing, revealing the author's feelings, whether positive, negative, or neutral, about a particular subject.}
\end{itemize}
The specific prompt is as follows,
\begin{quote}
    \setlength{\leftskip}{0.3cm}
    \setlength{\rightskip}{0.3cm}
\textit{You are a heavy social media user and are very familiar with the way of expression on the Internet. Analyze the following sentence, focusing on the content, emotional tone, implied meaning, and so on. Do nothing else. \{tweet}\}
\end{quote}

\subsection{Reasoning-Enhanced Debating Stage}
\subsubsection{Challenge:} The task of stance detection requires sophisticated reasoning. Authors often do not explicitly state their positions in a text. Instead, their stances may be implied through their sentiment towards certain entities or by mechanisms like comparison and contrast. Identifying these implicit stances requires detailed reasoning. Although large-scale language models possess some reasoning capabilities, their performance can be suboptimal in intricate reasoning tasks without proper guidance, which can affect the quality of stance detection results.
\subsubsection{Approach:} Drawing inspiration from recent works that leverage discussions or debates among large models to enhance their performance~\cite{du2023improving,chan2023chateval,liang2023encouraging}, especially in reasoning tasks, we introduce a reasoning-enhanced debating stage. In this stage, for every potential stance, an agent is designated. This agent seeks evidence from expert analyses of the text and advocates for its designated stance. Specifically, the inputs and outputs of agents in this stage are as follows.

\textbf{Input:} A text with a stance. The analyses of the text by the linguistic expert, the domain specialist, and the social media veteran.

\textbf{Output:} The debate from each agent for the stance they support, including the evidence it chooses and its logical chain.

The specific prompt is as follows,
\begin{quote}
    \setlength{\leftskip}{0.1cm}
\textit{Tweet:\{tweet\}. Linguistic analysis:\{LingResponse\}. The analysis of \{role\}:\{ExpertResponse\}. The analysis of a heavy social media user: \{UserResponse\}. You think the attitude behind the tweet is \{stance\} of \{target\}. Identify the top three pieces of evidence from the analyses that best support your opinion and argue for your opinion.}
\end{quote}
In our framework, we only engage in a single round of debate, reserving multi-round debates for future exploration. Directing agents to search for evidence and defend their aligned stances compels the large language model to establish logical connections between discerned textual features (as well as their multifaceted interpretations) and the actual underlying stance of the text. By having multiple agents debate in favor of different stances, the system encourages the large model's divergent thinking. These outputs subsequently feed into the stance conclusion stage, which renders a final, judicious judgment. 

\subsection{Stance Conclusion Stage}
To infer a conclusive stance from diverse agent debates, we introduce the stance conclusion stage. In this stage, a judger agent determines the final stance of a text based on both the text itself and the arguments presented by debater agents. The process is delineated as:

\textbf{Input:} A text with an embedded stance. Arguments from each agent, including evidence and their logical reasoning.

\textbf{Output:}The identified stance of the text.

The specific prompt can be as follows,
\begin{quote}
    \setlength{\leftskip}{0.2cm}
    \setlength{\rightskip}{0.2cm}
\textit{Determine whether the sentence is in favor of or against \{target\}, or is neutral. Sentence: \{tweet\}. Judge this in relation to the following arguments: Arguments that the attitude is in favor: \{FavorResponse\}. Arguments that the attitude is against: \{AgainstResponse\}. Arguments that the attitude is neutral: \{NeutralResponse\} Choose from: A: Against B: Favor C: Neutral}

\textit{Constraint: Answer with only the option above that is most accurate and nothing else.}
\end{quote}
The judger agent evaluates the text's inherent qualities, the evidence provided by debaters, and their logical frameworks to reach an informed decision. 

After going through the three stages mentioned above, we have effectively extracted the underlying stance towards the given target from the text.

\begin{table}[]
\centering
\begingroup
\resizebox{1\linewidth}{!}{
\begin{tabular}{c|c|cccc}
\hline
Dataset                & Target & Pro & Con & Neutral  \\ \hline
\multirow{6}{*}{SEM16} & DT     & 148 (20.9\%)   & 299 (42.3\%)     & 260 (36.8\%)            \\
                       & HC     & 163 (16.6\%)  & 565 (57.4\%)    & 256 (26.0\%)           \\
                       & FM     & 268 (28.2\%)  & 511 (53.8\%)    & 170 (17.9\%)            \\
                       & LA     & 167 (17.9\%)   & 544 (58.3\%)     & 222 (23.8\%)             \\
                       & A      & 124 (16.9\%)   & 464 (63.3\%)      & 145 (19.8\%)             \\
                       & CC     & 335 (59.4\%)   & 26 (4.6\%)      & 203 (36.0\%)             \\ \hline
\multirow{3}{*}{P-Stance} & Biden     & 3217 (44.1\%)  & 4079 (55.9\%)     & -       \\
                       & Sanders     & 3551 (56.1\%)   & 2774 (43.9\%)    & -        \\
                       & Trump    & 3663 (46.1\%)  & 4290 (53.9\%)    & -   \\ \hline
VAST                   & -      & 6952 (37.5\%)  & 7297 (39.3\%)    & 4296 (23.2\%)            \\ \hline
\end{tabular}}
\endgroup
\caption{Statistics of our utilized datasets.}
\label{tab:datasets}
\end{table}

\section{Experiments}\label{sec::experiments}
In this section, we describe the specific setup of our experiments.
\subsection{Datasets}
Following many existing works~\cite{liang2022zero,augenstein2016stance,li2023stance}, we conduct experiments on three widely-used datasets:

\textbf{SEM16}~\cite{mohammad2016semeval}. This dataset features six specific targets from diverse domains, namely \textit{Donald Trump} (DT), \textit{Hillary Clinton} (HC), \textit{Feminist Movement} (FM), \textit{Legalization of Abortion} (LA), \textit{Atheism} (A), and \textit{Climate Change is Real Concern} (CC). Each instance is classified into one of the three stance categories: \textit{Favor}, \textit{Against}, or \textit{None}.

\textbf{P-Stance}~\cite{li2021p}. This dataset focuses on the political domain, and comprises three targets: \textit{Donald Trump} (Trump), \textit{Joe Biden} (Biden), \textit{Bernie Sanders} (Sanders). Stance labels include \textit{Favor} and \textit{Against}. 

\textbf{VAST}~\cite{allaway2020zero}. This dataset is characterized by its large number of varying targets. An instance in VAST includes a sentence, a target, and a stance, which may be \textit{Pro}, \textit{Con}, or \textit{Neutral}.

The statistics of our utilized datasets are shown in Table~\ref{tab:datasets}. To ensure a fair comparison, We follow the majority of existing works~\cite{allaway2020zero,allaway2021adversarial,liang2022zero,zhang2019aspect} to test the performance of our model. Specifically, on the SEM16 and P-Stance datasets, we test the performance of our model on the test set. On VAST dataset, we test the performance of our model over zero-shot condition. To ensure a fair comparison with LLM-based baselines, we first sample the test set to replicate their results under their prompts, and then conduct experiments on the dataset. For zero-shot stance detection approaches, we evaluate their performance across all three datasets. However, for in-target stance detection methods, we assess their performance on SEM16 and P-Stance, because the targets within the VAST dataset are mainly few-shot or zero-shot. The datasets contain no personally identifiable information, but may contain offensive content because the text has a clear stance on topics such as religions, politics, climate, etc. We strictly adhere to the requirements of the respective licenses when using all datasets mentioned in the paper.
\begin{table*}[t]
\centering
\small
\begin{tabular}{c|cccccc|ccc|c}
\hline
\multirow{2}{*}{\textbf{Model}} & \multicolumn{6}{c|}{\textbf{SEM16(\%)}}                                                                                                                  & \multicolumn{3}{c|}{\textbf{P-Stance(\%)}}                                                             & \multicolumn{1}{c}{\textbf{VAST(\%)}}                                                               \\
                       & DT                             & HC                             & FM                & LA                & A                 & CC                & Trump                & Biden               &  Sanders                                                                         & All               \\ \hline
TOAD                   & 49.5                           & 51.2                           & 54.1              & 46.2              & 46.1              & 30.9              &  53.0             &  68.4          &  62.9        & 41.0                                                                          \\
TGA Net                & 40.7                           & 49.3                           & 46.6              & 45.2              & 52.7              & 36.6              &   -            & -              & -              &65.7           \\
BERT-GCN               & 42.3                           & 50.0                           & 44.3              & 44.2              & 53.6              & 35.5              & -             & -              & -              & 68.6                      \\
PT-HCL                 & 50.1                           & 54.5                           & \underline{54.6}              & 50.9              & \underline{56.5}              & 38.9              & -              & -              & -             & 71.6                      \\

JointCL &50.5&54.8&53.8&49.5&54.5&\underline{39.7}&62.0&59.0&73.0&\underline{72.3}\\
GPT-3.5                & 62.5                           & 68.7                          & 44.7              & 51.5              & 9.1               & 31.1              & 62.9            &80.0              &71.5                                  & 62.3              \\
GPT-3.5+COT            & \underline{63.3}                           & \underline{70.9}                           & 47.7              & \underline{53.4}              & 13.3              & 34.0              &  \underline{63.9}             & \underline{81.2}             & \underline{73.2}                                    & 68.9              \\
COLA(ours)             & \textbf{68.5} & $\textbf{81.7}^*$ & $\textbf{63.4}^*$ & $\textbf{71.0}^*$ & $\textbf{70.8}^*$ & $\textbf{65.5}^*$ & $\textbf{86.6}^*$ & \textbf{84.0} & $\textbf{79.7}^*$ & \textbf{73.0}   \\ \hline
\end{tabular}
\caption{Comparison of COLA and baselines in zero-shot stance detection task. Bold and underline refer to the best and 2nd-best performance. * denotes COLA improves the best baseline at $p < 0.05$ with paired t-test.}
\label{tab:zero-shot}
\end{table*}

\begin{table*}[t]
\centering
\small
\begin{tabular}{c|c|cccccc|ccc}
\hline
\multirow{2}{*}{\textbf{Category}} &
  \multirow{2}{*}{\textbf{Model}} &
  \multicolumn{6}{c|}{\textbf{SEM16(\%)}} &
  \multicolumn{3}{c}{\textbf{P-Stance(\%)}} \\
                       &          & DT   & HC   & FM   & LA   & A    & CC   & Trump   & Biden   & Sanders      \\ \hline
                       & BiCond   & 59.0 & 56.1 & 52.9 & 61.2 & 55.3 & 35.6  & 73.0 & 69.4 & 64.6 \\
                       & BERT     & 57.9 & 61.3 & 59.0 & 63.1 & 60.7 & 38.8  & 67.7 & 73.1 & 68.2 \\
In-target Labeled Data & CrossNet & 60.2 & 60.2 & 55.7 & 61.3 & 56.4 & 40.1  & 58.0 & 65.0 & 53.0 \\
Dependent Methods      & ATT-LSTM & 55.3 & 59.8 & 55.3 & 62.6 & 55.9 & 39.2  & - & - & - \\
                       & ASGCN    & 58.7 & 61.0 & 58.7 & 63.2 & 59.5 & 40.6  & \underline{77.0} & \underline{78.4} & 70.8 \\
 &
  TPDG &
  \underline{63.0} &
  \underline{73.4} &
  \textbf{67.3} &
  \textbf{74.7} &
  \underline{64.7} &
  \underline{42.3} &
  76.8 &
  78.1 &
  \underline{71.0}  \\ \hline
Zero-shot Method &
  COLA(ours) &
  \textbf{68.5} &
  $\textbf{81.7}^*$ &
  \underline{63.4} &
  \underline{71.0} &
  \textbf{70.8} &
  $\textbf{67.5}^*$ &
  $\textbf{86.6}^*$ &
  $\textbf{84.0}^*$ &
  $\textbf{79.7}^*$ \\ \hline
\end{tabular}
\caption{Comparison of zero-shot COLA and baselines fully trained on labeled data for the in-target stance detection task. Bold and underline refer to the best and 2nd-best performance. * denotes COLA improves the best baseline at $p < 0.05$ with paired t-test.}
\label{tab:in-target}
\end{table*}
\subsection{Implementation Details}

\subsubsection{Implementation of COLA}
In our study, we employ the GPT-3.5 Turbo model, provided by OpenAI, as our backbone. We opt for GPT-3.5 Turbo primarily due to its superior performance, cost-effectiveness, and the ease of interaction offered via OpenAI API. These attributes not only facilitate efficient research but also ensure the usability of our methodology for future application. By utilizing the system instruction feature available through OpenAI API, we instruct the model to act as various agent roles, feeding text inputs via prompts and collecting textual outputs from the model. To maximize reproducibility, we set the temperature parameter to 0. The reported results are the average of 5 repeated runs to ensure statistical reliability.
\footnote{The source code of our proposed framework is released at https://github.com/tsinghua-fib-lab/COLA}. 

\subsubsection{Evaluation Metric}
For SEM16 and P-Stance datasets, following  previous works~\cite{allaway2021adversarial,li2023stance}, we calculate $F_{avg}$, which represents the average of F1 scores for \textit{Favor} and \textit{Against}. For the VAST dataset, we adopt the commonly-used method from Allaway et al. (\citeyear{allaway2020zero}) and compute the Macro-F1 score to assess model performance.

\subsection{Comparison Methods}

We compare COLA with state-of-the-art (SOTA) methods in stance detection. We conduct comparisons with methods for two tasks: zero-shot stance detection and in-target stance detection. 

We compare our method with various zero-shot stance detection methods. This includes adversarial learning method: TOAD~\cite{allaway2021adversarial}, contrastive learning methods: PT-HCL~\cite{liang2022zero}, JointCL~\cite{liang2022jointcl}, Bert-based techniques: TGA-Net~\cite{allaway2020zero} and Bert-GCN~\cite{liu2021enhancing}. We also include two baselines based on large language models: GPT-3.5 Turbo and GPT-3.5 Turbo+Chain-of-thought(COT), both of which can be considered zero-shots, implemented in strict accordance with Zhang et al. (\citeyear{zhang2022would}) and Zhang et al. (\citeyear{zhang2023investigating}), respectively.

To further verify the performance of our model, we compare our model to in-target stance detection methods. Such methods undergo extensive training on datasets for a given target and are then evaluated on the test set of the same target. In contrast, our method remains strictly zero-shot, with \textbf{no fine-tuning} applied to our backbone model. We compare our approach with various in-target stance detection baselines, including RNN-based methods: BiCond~\cite{augenstein2016stance}, and ATT-LSTM~\cite{wang2016attention}; Attention-based method: CrossNet~\cite{xu2018cross}; Bert-based method: BERT~\cite{devlin2018bert}; and Graph-based methods: ASGCN~\cite{zhang2019aspect} and TPDG~\cite{liang2021target}. 

For non-LLM approaches, we retrieve results from existing literature for a comprehensive comparison~\cite{allaway2020zero,allaway2021adversarial,liu2021enhancing,liang2021target,liang2022zero,huang2023knowledge,khiabani2024socialpet}.

\section{Results and Discussions}\label{sec::results}
In this section, we aim to answer the following research questions (RQs) with the help of experimental results:

RQ1: How is the performance of COLA compared with state-of-the-art stance detection models? \textbf{(Accuracy)}

RQ2: Is every component in our model effective and contributory to performance enhancement? \textbf{(Effectiveness)}

RQ3: Can our model explain the rationale and logic behind its stance determinations? \textbf{(Explainability)}

RQ4: Is our framework adaptable to other text classification tasks related to web and social media content analysis? \textbf{(Versatility)}
\subsection{Overall Performance (RQ1)}

In Table~\ref{tab:zero-shot}, we present the zero-shot stance detection performance of COLA across three datasets in comparison to baseline methods. Furthermore, Table~\ref{tab:in-target} showcases the results of both our zero-shot COLA and the in-target labeled data dependent baselines on the SEM16 and P-Stance datasets for the in-target stance detection task. Overall results have demonstrated the strong performance of our approach. Specifically, the key findings are enumerated below.
\begin{itemize}

\item{\textbf{Our method outperforms the state-of-the-art zero-shot stance detection approaches across all metrics.} On most metrics across three datasets, our model demonstrates statistically significant improvements over the best baseline. For the CC and LA targets in the SEM16 dataset, our approach achieves substantial gains over the best baseline, with absolute increases in $F_{avg}$ of 16.9\% and 26.6\% respectively. On the VAST dataset, which comprises tens of thousands of instances, our model secures a notable absolute boost of 0.7\% in the overall Macro-F1 Score. This attests to the robust zero-shot stance detection capabilities of our approach.}
\item{\textbf{The performance of our approach matches that of in-target stance detection baselines.} The zero-shot stance detection performance of our method is closely aligned with that of the state-of-the-art in-target stance detection techniques, even when they are fully trained on corresponding targets. On the SEM16 dataset, our approach significantly outperforms the best baseline, TPDG, on the HC and CC targets, while maintaining comparable performance on other targets. On the P-Stance dataset, our method consistently outperforms the performance of all baselines across all targets. Remarkably, even though these comparison methods have been extensively trained on their respective targets, our approach still sustains comparable or superior performance, underscoring our method's strong performance.}
\item{\textbf{Direct application of LLMs may yield poor performance, especially on abstract concept targets.} On the SEM16 dataset, for the targets A (\textit{Atheism}) and CC (\textit{Climate Change is a Real Concern}), GPT-3.5 achieves only 9.1\% and 31.1\% in $F_{avg}$ respectively. Even with the enhanced GPT-3.5+COT, the scores are merely 13.3\% and 34.0\%. Across almost all datasets and metrics, the performance of simply deploying large language models significantly lags behind our proposed method. This underscores the limitations of directly using large language models for stance detection tasks, especially in handling stances towards abstract concept targets, highlighting the necessity and validity of our design.}
\end{itemize}

\begin{table}[]

\centering
\begingroup\small
\resizebox{1.0\linewidth}{!}{
\begin{tabular}{c|cccccc}
\hline
\multirow{2}{*}{\textbf{Model}} & \multicolumn{6}{c}{\textbf{SEM16(\%)}}  \\
                                & DT   & HC   & FM   & LA   & A    & CC   \\ \hline
Flan-UL2                        & 64.4 & 70.1 & 65.3 & 67.3 & 57.5 & 68.5 \\
Flan-UL2 with COLA     & \textbf{64.9} & \textbf{72.3} & \textbf{65.7} & \textbf{69.8} & \textbf{61.6} & \textbf{75.1} \\
ChatGLM-2 6B                    & 37.9 & 60.2 & 42.0 & 43.2 & 41.0 & 13.7 \\
ChatGLM-2 6B with COLA & \textbf{45.3} & \textbf{60.6} & \textbf{55.4} & \textbf{43.9} & \textbf{43.6} & \textbf{37.6} \\ \hline
\end{tabular}}
\endgroup
\caption{Performance of COLA when utilizing Flan-UL2 or GhatGLM-2 6B as backbones.}
\label{tab:FlanGLM}
\end{table}

To confirm that our method can enhance stance detection based on LLMs and not just augment the capabilities of the closed-source GPT-3.5 Turbo, we conduct experiments using other LLM backbones. Specifically, we utilized the Flan-UL2 and ChatGLM2-6B models for experiments on the SEM16 dataset. Flan-UL2 demonstrates notable performance in stance detection tasks~\cite{ziems2023can}, while ChatGLM2-6B is a more commonly employed model. The results of these experiments are presented in Table~\ref{tab:FlanGLM}.

It can be observed that the performance of Flan-UL2 surpassed that of GPT-3.5 Turbo, while ChatGLM2 6B significantly underperforms in comparison. On the SEM16 dataset, regardless of whether the LLM backbone is Flan-UL2 or ChatGLM2-6B, the performance of COLA consistently exceeded that of the LLM backbones. Notably, on the less efficient ChatGLM2-6B, COLA contributes to a more significant performance enhancement, exemplified by a 23.9\% absolute increase in $F_{avg}$ on the CC Target and a 13.4\% absolute increase in $F_{avg}$ on FM. These experimental results demonstrate that our method can enhance stance detection performance not only for GPT-3.5 Turbo but also for other LLMs.

\subsection{Ablation Study (RQ2)}
To investigate the impacts of each module in our design, we conduct ablation studies to assess the performance of our framework when each module is removed. The results are shown in Table~\ref{tab:ablation}, which demonstrate that every module in our framework contributes to performance enhancement. In the following, we provide a detailed description of the results.
\subsubsection{Study on multidimensional text analysis stage.}
During the multidimensional text analysis stage, three expert agents from different domains concurrently analyze the text. We individually remove each of these experts to assess the performance of our approach. We also evaluated the performance when all expert analyses are excluded. The results show that the removal of any expert agent results in a certain degree of performance degradation in all cases except for FM on SEM16 dataset. Moreover, eliminating the entire multidimensional text analysis stage leads to a significant performance drop. The most pronounced performance decline is observed for the LA target on SEM16 dataset. Removing the Linguistic Expert, Domain Specialist, and Social Media Veteran leads to decreases in $F_{avg}$ to 68.9\%, 67.9\%, and 64.1\%, respectively. What's more, without the multidimensional text analysis stage, the $F_{avg}$ drops to a mere 63.8\%. This could be attributed to the complexity of the LA topic across various domains such as religions and society. These findings underscore the effectiveness of our multidimensional text analysis stage and the design of each agent therein.
\subsubsection{Study on reasoning-enhanced debating stage.}
In the reasoning-enhanced debating phase, we introduce debates among agents with differing perspectives to augment the reasoning capabilities of our LLM-based system. We remove this stage and let the judger agent directly deduce the text's stance from the expert agents' text analysis, aiming to verify the effectiveness of the debating design. Removing the debating stage results in a greater performance loss than removing the text analysis stage. Upon the removal of the debating stage, our method experiences a noticeable performance degradation. The most significant drops are observed for the abstract concept targets LA (\textit{Legalization of Abortion}), CC (\textit{Climate Change is Real Concern}) and A (\textit{Atheism}), with the absolute $F_{avg}$ declining by 31.2\%, 14.1\%, and 11.2\%, respectively. This indicates that the reasoning-enhanced debating stage offers substantial benefits, especially when dealing with relatively abstract targets. The results validate the effectiveness of the reasoning-enhanced debating stage design.

In summary, comprehensive ablation studies have demonstrated the effectiveness of each module in our designed method.
\begin{table}[t]
\centering
\begingroup\small
\resizebox{1.0\linewidth}{!}{
\begin{tabular}{c|cccccc}
\hline
\multirow{2}{*}{\textbf{Model}} & \multicolumn{6}{c}{\textbf{SEM16(\%)}}                                                                                                                                                                            \\
                       & DT                             & HC                             & FM                & LA                & A                 & CC                         \\ \hline
COLA                            & 68.5 &	81.7&	63.4&	71.0&	70.8&	67.5 \\
w/o Liguisitic Expert           & 64.3&	80.5&	63.3&	68.9&	69.9&	65.5 \\
w/o Domain Specialist           &66.5&	79.2&	64.4&	67.9&	70.7&	65.4 \\
w/o Social Media Veteran           & 64.8&	76.8&	64.5&	64.1&	67.7&	63.5 \\
w/o Text Analysis Stage               & 64.4&	77.2&	65.7&	63.8&	67.0&	62.3 \\
w/o Debating Stage                    & 64.7&	74.9&	62.5&	39.2&	59.6&	53.4 \\ \hline
\end{tabular}}
\endgroup
\caption{Experimental results of ablation study.}
\label{tab:ablation}
\end{table}

\begin{figure}[ht]
    \centering
    \includegraphics[width=0.95\linewidth]{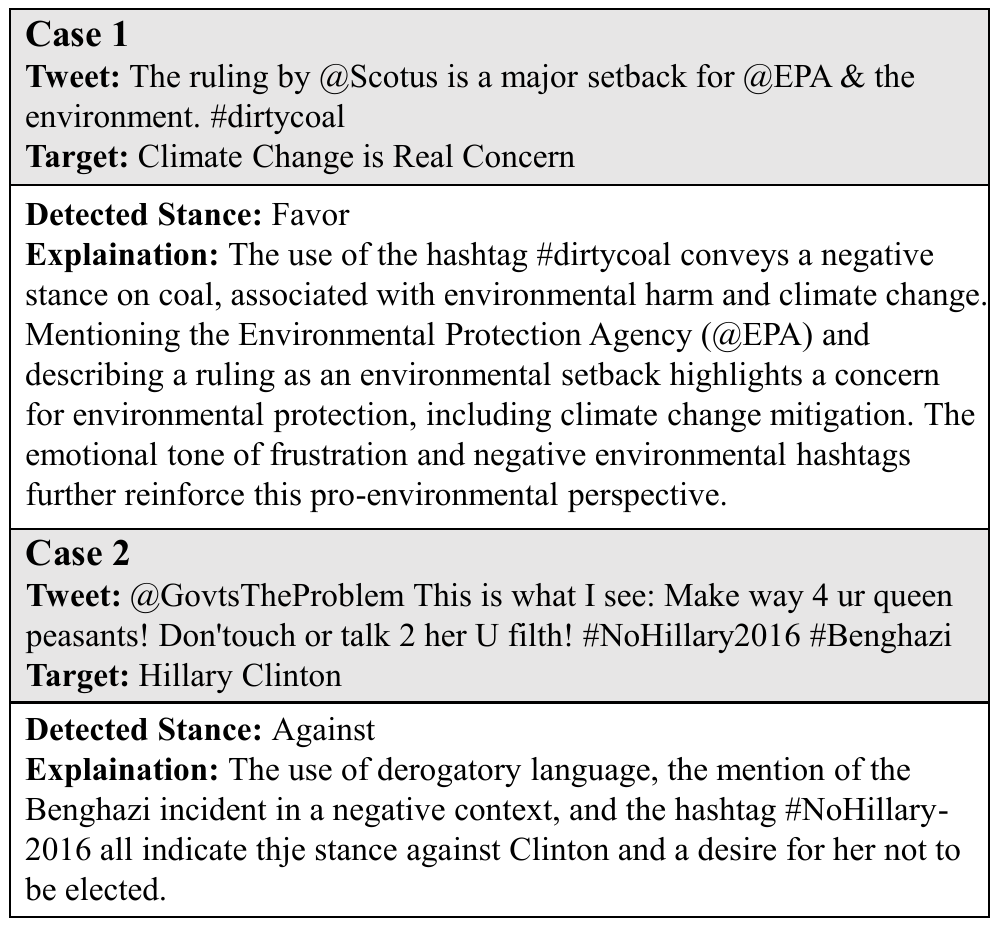}
    \caption{Cases of explainations generated by our approach.}
    \label{fig:explain}
\end{figure}

\begin{table}[]
\centering
\begingroup\small
\resizebox{1.0\linewidth}{!}{
\begin{tabular}{c|cccccc}
\hline
\multirow{2}{*}{\textbf{Method}} & \multicolumn{6}{c}{\textbf{SEM16(\%)}}                                                        \\
                                 & DT            & HC            & FM            & LA            & A             & CC            \\ \hline
GPT-3.5                          & 69.0          & 74.0          & 59.1         & 52.0         & 8.1          & 24.7          \\
COLA                             & \textbf{71.2} & 75.9          & 69.1          & \textbf{71.0} & \textbf{62.3} & \textbf{64.0} \\
GPT-3.5+COLA's Explainations      & 69.4          & \textbf{77.7} & \textbf{70.7} & 66.7          & 61.9          & 54.5          \\ \hline
\end{tabular}}
\endgroup
\caption{Performance of GPT-3.5 Turbo, COLA and GPT-3.5 Turbo with explainations generated by COLA. Experiments are conducted on the whole SEM16 dataset. Best scores are in bold.}
\vspace{-0.4cm}
\label{tab:explainable}
\end{table}

\begin{table*}[]
\centering
\small
\begin{tabular}{c|c|cc|cc|cc}
\hline
\multirow{2}{*}{\textbf{Category}} &
  \multirow{2}{*}{\textbf{Model}} &
  \multicolumn{2}{c|}{\textbf{Restaurant14(\%)}} &
  \multicolumn{2}{c|}{\textbf{Laptop(\%)}} &
  \multicolumn{2}{c}{\textbf{Restaurant15(\%)}} \\
                                   &               & Accuracy      & Macro-F1      & Accuracy      & Macro-F1      & Accuracy      & Macro-F1      \\ \hline
Labeled Data                       & DGEDT         & \textbf{86.3}         & 80.0          & 79.8          &  75.6        & 84.0          & 71.0          \\
Dependent Methods                  & dotGCN        & 86.2 & \textbf{80.5} & 81.0          & \textbf{78.1} & 85.2          & \textbf{72.7}          \\ \hline
\multirow{2}{*}{Zero-shot Methods} & GPT-3.5 Turbo & 70.6          & 59.7          & 85.0         & 66.7          & 84.0          & 62.4          \\
                                   & Ours          & 74.1          & 65.7          & \textbf{87.0} & 67.5          & \textbf{90.5} & 64.3 \\ \hline
\end{tabular}
\caption{Performance of our framework and baselines on aspect-based sentiment analysis. Best scores are in bold.}
\label{tab:sentiment}
\end{table*}

\begin{table}[]
\centering
\small
\begin{tabular}{ccc}
\hline
\textbf{Model} & \textbf{Accuracy(\%)} & \textbf{F1-Score(\%)} \\ \hline
Hybrid RCNN    & 74.8                  & 59.6                  \\
GPT-3.5 Turbo  & 67.6                  & 56.0                  \\
Ours           & \textbf{76.5}                  & \textbf{63.9}                  \\ \hline
\end{tabular}
\caption{Performance of our framework and baselines on persuasion prediction. Best scores are in bold.}
\label{tab:persuasion}
\end{table}

\subsection{Study on Explainablity (RQ3)}
An explainable artificial intelligence (XAI) is one that offers clear insights or justifications to make its decisions comprehensible~\cite{arrieta2020explainable}. By elucidating its decision-making processes, an XAI augments transparency and reinforces model trustability~\cite{das2020opportunities}. Large language models inherently possess the capability to explain their outputs. By prompting them about the rationale behind their decisions, we can obtain explanations for their determinations directly. To delve deeper into the explainablility of our approach, we conduct both case studies and quantitative experiments to verify its ability to generate clear and reasonable explanations.
\subsubsection{Case Studies.}
During the stance conclusion stage, we mandate the judger agent to provide outputs in a JSON format, consisting of two components: the stance and a concise explanation not exceeding 100 tokens. We conduct our experiments on the SEM16 dataset. After closely examining the generated outputs, we find that our model can provide clear explanations for its decisions. In Figure~\ref{fig:explain}, we show two cases to illustrate, which are discussed as follows:
\begin{itemize}
    \item In the first case, the tweet \textit{``The ruling by @Scotus is a major setback for @EPA \& the environment. \#dirtycoal''} agrees that climate change is a real concern. Our model detects this stance. In its generated explanation, the model discerns the mention of the EPA and the usage of the \#dirtycoal tag, indicating an environmental concern. Moreover, the model perceives an emotional tone of frustration, further reflecting a pro-environmental perspective. 
    \item In the second case, the tweet \textit{``@GovtsTheProblem This is what I see: Make way 4 ur queen peasants! Don't touch or talk 2 her U filth! \#NoHillary2016 \#Benghazi''} portrays an opposing stance toward Hillary. Our model rationally explains its judgment from a linguistic perspective (utilization of derogatory language), a domain-specialist perspective (mentioning the Benghazi incident in a negative context), and a social media lens (the hashtag \#NoHillary2016). These cases validate the model's proficiency in generating clear and reasonable explanations.
\end{itemize}

\subsubsection{Quantitative Experiments.}
To further validate our model's ability to produce clear and logical explanations, we conduct quantitative experiments. For the SEM16 dataset, we collect explanations (from the second part of the JSON output) related to each instance's stance generated by COLA. These explanations, along with the original text, are fed into the GPT-3.5 Turbo model. We inform the model that these explanations could be used as references for its decisions. As a result, we obtain a new set of judgments from the model. It's evident that the performance of GPT-3.5 Turbo significantly improves by incorporating explanations generated by COLA in addition to the original texts, as presented in Table~\ref{tab:explainable}. Note that we do experiments on the whole SEM16 dataset here, rather than the test set, to enhance the credibility of the results. There is a noticeable increase for the A(\textit{Atheism}) and CC(\textit{Climate Change is Real Concern}) targets, with $F_{avg}$ improving by 51.6 and 29.3 points, respectively. For the HC(\textit{Hillary Clinton}) and FM(\textit{Feminist Movement}) targets, the results even exceed that of COLA. This further confirms our model's strong ability in generating clear and logical explanations.

Overall, both case studies and quantitative experiments have demonstrated the high explainability of our method. Its high explainability and accuracy make it a trustworthy approach.

\subsection{Study on Versatility (RQ4)}
Our proposed COLA can be summarized as an Analyst-Debater-Summarizer framework. In this section, we conduct experiments to validate that the Analyst-Debater-Summarizer framework can be applied to other text classification tasks for text analysis on web and social media, not just as an ad-hoc approach for stance detection. We perform experiments on two additional text classification tasks: aspect-based sentiment analysis and persuasion prediction. We select aspect-based sentiment analysis because it demands precise understanding of sentiments tied to specific elements in text, reflecting the detailed analysis capability of our framework. Meanwhile, persuasion prediction is chosen due to its emphasis on detecting underlying intent, highlighting COLA's ability to adeptly handle intricate conversational dynamics commonly seen in web and social media exchanges.

\subsubsection{Aspect-based Sentiment Analysis} 
\begin{itemize}
    \item \textbf{Experimental Setup:} Aspect-based sentiment analysis is to determine the sentiment polarity (\textit{Positive}, \textit{Negative}, or \textit{Neutral}) expressed towards each aspect mentioned in the text ~\cite{pontiki-etal-2014-semeval}. In this task, we modify the debater component in our original framework to engage in sentiment debates instead of stance debates, while keeping other design unchanged. We evaluate our approach's performance on the Restaurant14, Restaurant 15, and Laptop datasets from SemEval14~\cite{pontiki-etal-2014-semeval} and SemEval15~\cite{pontiki2016semeval}. We follow Chen et al. (\citeyear{chen2017recurrent}) and use Accuracy and Macro-F1 score as evaluation metrics. We compare our approach with state-of-the-art models that require training, namely DGEDT~\cite{tang2020dependency} and dotGCN~\cite{chen2022discrete}.
    \item \textbf{Results:} The experimental results are presented in Table~\ref{tab:sentiment}. It can be observed that our zero-shot method performs comparably to the best baseline models that rely on labeled data. On the Restaurant15 dataset, our approach even outperforms the top baseline on Accuracy. Another crucial finding is that our approach consistently outperforms directly applying GPT-3.5 Turbo while maintaining ease of use.
\end{itemize}

\subsubsection{Persuasion Prediction} 
\begin{itemize}
    \item \textbf{Experimental Setup:} Following Ziems et al. (\citeyear{ziems2023can}), we define persuasion prediction as determining whether one party in a conversation is persuaded after the conversation ends. In this task, we replace the three experts in our original framework with two experts: a domain expert and a psychologist. They provide detailed analyses of various concepts and nouns in the conversation topics and analyze the psychological changes of the individuals involved. The debaters are modified to argue for whether a participant in the conversation has been persuaded. We use the dataset provided by Wang et al.(\citeyear{wang2019persuasion}) and follow their evaluation metrics, using Accuracy and Macro-F1.
   \item \textbf{Results:} We compare our approach with Hybrid RCNN~\cite{wang2019persuasion} and GPT-3.5 Turbo, and the results are presented in Table~\ref{tab:persuasion}. The experimental results show that our approach achieves better performance compared to the baseline and a significant improvement over GPT-3.5 Turbo.
\end{itemize}
The Analyst-Debater-Summarizer framework has proven to be highly successful in both aspect-based sentiment analysis and persuasion classification tasks. On a series of tasks, our zero-shot framework performs on par with state-of-the-art baselines that rely on training data and significantly outperforms direct application of GPT-3.5 Turbo. These experiments demonstrate the versatility of our approach.

\subsection{Discussions}
In the aforementioned experiment, we extensively evaluate the performance of our approach across various dimensions, which are listed as follows:
\begin{itemize}
\item First, from the perspective of our method's design rationale, the ablation study confirms that every component in our approach contributes to a performance boost, indicating that the design is free of redundancy and can be considered efficacious.
\item Second, in comparison with existing methods, experimental evidence shows that our approach outperforms all other zero-shot methods on stance detection. Furthermore, its performance is on par with in-target stance detection methods that rely on in-target labeled data, exhibiting impressive accuracy.
\item In addition, for two other text classification tasks related to web and social media content analysis, our method achieves results comparable to state-of-the-art baselines, underscoring its versatility.
\item What's more, from a practical application standpoint, our method does not require additional training for the model. Instead, it can be implemented by interacting with existing large language models through APIs or other means, showcasing its strong usability.
\item Finally, the experiments also prove that our framework can provide clear and rational explanations for its decisions, ensuring a high degree of explainability. Such generated explanations can bolster users' trust in our approach and are conducive to further analysis. 
\end{itemize}   
Given these advantages, our method promises a broad range of applications.

\section{Conclusion and Future Work}\label{sec::conclusion}
In this work, we harness the strong capabilities of LLMs for advanced stance detection. We propose COLA, where multiple LLM-based agents collaborate to reach an conclusion. This method encompasses three stages: the multidimensional text analysis stage, the reasoning-enhanced debating stage, and the stance conclusion stage. Experimental results demonstrate that our approach achieves high accuracy, effectiveness, explainability, and versatility, showcasing its significant applicability.

Due to the absence of real-time training data for large language models, the performance in analyzing real-time topics might be slightly compromised. For future work, we intend to incorporate a real-time updating knowledge base into the text analysis stage to enhance our framework's capability to analyze texts that include current events. We plan to first retrieve relevant information from the real-time knowledge base, and then have the LLMs use this information to generate analytical texts. Furthermore, there remains vast potential for exploring its implementation in addressing extensive text analysis tasks on web and social media.

\section{Ethics Statement}
All the datasets that we utilize for this research are open-access datasets. The VAST dataset provides full text data directly. In accordance with Twitter's privacy agreement for academic purposes, the SEM16 and P-Stance datasets are accessed using the official Twitter API\footnote{https://developer.twitter.com/en/docs/twitter-api} to retrieve complete text data based on Tweet IDs. The datasets do not include any personally identifiable information, but they might include offensive content as the text expresses strong opinions on subjects like religion, politics, climate, etc. We consistently comply with the respective licenses' requirements when utilizing all the datasets referenced in the paper. We use the GPT-3.5 Turbo API service provided by OpenAI, with adherence to OpenAI's terms and policies. 

In our primary experiments, we employed GPT-3.5 Turbo as the backbone. While the use of closed-source LLMs entails significant financial costs, our framework has demonstrated improved stance detection performance with open-source LLMs as well. It is important to acknowledge that running LLMs requires substantial energy, a common issue for all algorithms based on LLMs. We look forward to advancements in energy-efficient hardware technologies that could alleviate this concern.

Regarding potential misuse, we recognize that our technology, like many others, carries the risk of being exploited for unethical purposes, such as such as for silencing critics or identifying and targeting dissenting voices on social media by certain entities. We urge users of our technology to commit to responsible and ethical usage. It is crucial to balance technological advancement with a conscientious approach to mitigate risks, especially in areas like stance detection that intersect with sensitive societal and political domains.

\section{Acknowledgements}
This work is supported by the National Science Foundation of China under U23B2030, 62272262 and 72342032.

\bibliography{aaai23}

\begin{thebibliography}{54}
\providecommand{\natexlab}[1]{#1}

\bibitem[{AlDayel and Magdy(2021)}]{aldayel2021stance}
AlDayel, A.; and Magdy, W. 2021.
\newblock Stance detection on social media: State of the art and trends.
\newblock \emph{Information Processing \& Management}, 58(4): 102597.

\bibitem[{Allaway and McKeown(2020)}]{allaway2020zero}
Allaway, E.; and McKeown, K. 2020.
\newblock Zero-shot stance detection: A dataset and model using generalized topic representations.
\newblock \emph{arXiv preprint arXiv:2010.03640}.

\bibitem[{Allaway, Srikanth, and McKeown(2021)}]{allaway2021adversarial}
Allaway, E.; Srikanth, M.; and McKeown, K. 2021.
\newblock Adversarial learning for zero-shot stance detection on social media.
\newblock \emph{arXiv preprint arXiv:2105.06603}.

\bibitem[{Arrieta et~al.(2020)Arrieta, D{\'\i}az-Rodr{\'\i}guez, Del~Ser, Bennetot, Tabik, Barbado, Garc{\'\i}a, Gil-L{\'o}pez, Molina, Benjamins et~al.}]{arrieta2020explainable}
Arrieta, A.~B.; D{\'\i}az-Rodr{\'\i}guez, N.; Del~Ser, J.; Bennetot, A.; Tabik, S.; Barbado, A.; Garc{\'\i}a, S.; Gil-L{\'o}pez, S.; Molina, D.; Benjamins, R.; et~al. 2020.
\newblock Explainable Artificial Intelligence (XAI): Concepts, taxonomies, opportunities and challenges toward responsible AI.
\newblock \emph{Information fusion}, 58: 82--115.

\bibitem[{Augenstein et~al.(2016)Augenstein, Rockt{\"a}schel, Vlachos, and Bontcheva}]{augenstein2016stance}
Augenstein, I.; Rockt{\"a}schel, T.; Vlachos, A.; and Bontcheva, K. 2016.
\newblock Stance detection with bidirectional conditional encoding.
\newblock \emph{arXiv preprint arXiv:1606.05464}.

\bibitem[{Bar-Haim et~al.(2017)Bar-Haim, Bhattacharya, Dinuzzo, Saha, and Slonim}]{bar2017stance}
Bar-Haim, R.; Bhattacharya, I.; Dinuzzo, F.; Saha, A.; and Slonim, N. 2017.
\newblock Stance classification of context-dependent claims.
\newblock In \emph{Proceedings of the 15th Conference of the European Chapter of the Association for Computational Linguistics: Volume 1, Long Papers}, 251--261.

\bibitem[{Brown et~al.(2020)Brown, Mann, Ryder, Subbiah, Kaplan, Dhariwal, Neelakantan, Shyam, Sastry, Askell et~al.}]{brown2020language}
Brown, T.; Mann, B.; Ryder, N.; Subbiah, M.; Kaplan, J.~D.; Dhariwal, P.; Neelakantan, A.; Shyam, P.; Sastry, G.; Askell, A.; et~al. 2020.
\newblock Language models are few-shot learners.
\newblock \emph{Advances in neural information processing systems}, 33: 1877--1901.

\bibitem[{Cai et~al.(2023)Cai, Wang, Ma, Chen, and Zhou}]{cai2023large}
Cai, T.; Wang, X.; Ma, T.; Chen, X.; and Zhou, D. 2023.
\newblock Large language models as tool makers.
\newblock \emph{arXiv preprint arXiv:2305.17126}.

\bibitem[{Chan et~al.(2023)Chan, Chen, Su, Yu, Xue, Zhang, Fu, and Liu}]{chan2023chateval}
Chan, C.-M.; Chen, W.; Su, Y.; Yu, J.; Xue, W.; Zhang, S.; Fu, J.; and Liu, Z. 2023.
\newblock ChatEval: Towards Better LLM-based Evaluators through Multi-Agent Debate.
\newblock \emph{arXiv preprint arXiv:2308.07201}.

\bibitem[{Chen et~al.(2022)Chen, Teng, Wang, and Zhang}]{chen2022discrete}
Chen, C.; Teng, Z.; Wang, Z.; and Zhang, Y. 2022.
\newblock Discrete opinion tree induction for aspect-based sentiment analysis.
\newblock In \emph{Proceedings of the 60th Annual Meeting of the Association for Computational Linguistics (Volume 1: Long Papers)}, 2051--2064.

\bibitem[{Chen et~al.(2017)Chen, Sun, Bing, and Yang}]{chen2017recurrent}
Chen, P.; Sun, Z.; Bing, L.; and Yang, W. 2017.
\newblock Recurrent attention network on memory for aspect sentiment analysis.
\newblock In \emph{Proceedings of the 2017 conference on empirical methods in natural language processing}, 452--461.

\bibitem[{Das and Rad(2020)}]{das2020opportunities}
Das, A.; and Rad, P. 2020.
\newblock Opportunities and challenges in explainable artificial intelligence (xai): A survey.
\newblock \emph{arXiv preprint arXiv:2006.11371}.

\bibitem[{Devlin et~al.(2018)Devlin, Chang, Lee, and Toutanova}]{devlin2018bert}
Devlin, J.; Chang, M.-W.; Lee, K.; and Toutanova, K. 2018.
\newblock Bert: Pre-training of deep bidirectional transformers for language understanding.
\newblock \emph{arXiv preprint arXiv:1810.04805}.

\bibitem[{Du et~al.(2023)Du, Li, Torralba, Tenenbaum, and Mordatch}]{du2023improving}
Du, Y.; Li, S.; Torralba, A.; Tenenbaum, J.~B.; and Mordatch, I. 2023.
\newblock Improving Factuality and Reasoning in Language Models through Multiagent Debate.
\newblock \emph{arXiv preprint arXiv:2305.14325}.

\bibitem[{Gao et~al.(2023{\natexlab{a}})Gao, Lan, Li, Yuan, Ding, Zhou, Xu, and Li}]{gao2023large}
Gao, C.; Lan, X.; Li, N.; Yuan, Y.; Ding, J.; Zhou, Z.; Xu, F.; and Li, Y. 2023{\natexlab{a}}.
\newblock Large language models empowered agent-based modeling and simulation: A survey and perspectives.
\newblock \emph{arXiv preprint arXiv:2312.11970}.

\bibitem[{Gao et~al.(2023{\natexlab{b}})Gao, Lan, Lu, Mao, Piao, Wang, Jin, and Li}]{gao2023s}
Gao, C.; Lan, X.; Lu, Z.; Mao, J.; Piao, J.; Wang, H.; Jin, D.; and Li, Y. 2023{\natexlab{b}}.
\newblock S$^{3}$: Social-network Simulation System with Large Language Model-Empowered Agents.
\newblock \emph{arXiv preprint arXiv:2307.14984}.

\bibitem[{Gr{\v{c}}ar et~al.(2017)Gr{\v{c}}ar, Cherepnalkoski, Mozeti{\v{c}}, and Kralj~Novak}]{grvcar2017stance}
Gr{\v{c}}ar, M.; Cherepnalkoski, D.; Mozeti{\v{c}}, I.; and Kralj~Novak, P. 2017.
\newblock Stance and influence of Twitter users regarding the Brexit referendum.
\newblock \emph{Computational social networks}, 4: 1--25.

\bibitem[{Hong et~al.(2023)Hong, Zheng, Chen, Cheng, Zhang, Wang, Yau, Lin, Zhou, Ran et~al.}]{hong2023metagpt}
Hong, S.; Zheng, X.; Chen, J.; Cheng, Y.; Zhang, C.; Wang, Z.; Yau, S. K.~S.; Lin, Z.; Zhou, L.; Ran, C.; et~al. 2023.
\newblock Metagpt: Meta programming for multi-agent collaborative framework.
\newblock \emph{arXiv preprint arXiv:2308.00352}.

\bibitem[{Huang et~al.(2023)Huang, Zhang, Li, Zhang, Sun, Luo, and Peng}]{huang2023knowledge}
Huang, H.; Zhang, B.; Li, Y.; Zhang, B.; Sun, Y.; Luo, C.; and Peng, C. 2023.
\newblock Knowledge-enhanced prompt-tuning for stance detection.
\newblock \emph{ACM Transactions on Asian and Low-Resource Language Information Processing}, 22(6): 1--20.

\bibitem[{Jang and Allan(2018)}]{jang2018explaining}
Jang, M.; and Allan, J. 2018.
\newblock Explaining controversy on social media via stance summarization.
\newblock In \emph{The 41st International ACM SIGIR Conference on Research \& Development in Information Retrieval}, 1221--1224.

\bibitem[{Khiabani and Zubiaga(2024)}]{khiabani2024socialpet}
Khiabani, P.~J.; and Zubiaga, A. 2024.
\newblock SocialPET: Socially Informed Pattern Exploiting Training for Few-Shot Stance Detection in Social Media.
\newblock \emph{arXiv preprint arXiv:2403.05216}.

\bibitem[{K{\"u}{\c{c}}{\"u}k and Can(2020)}]{kuccuk2020stance}
K{\"u}{\c{c}}{\"u}k, D.; and Can, F. 2020.
\newblock Stance detection: A survey.
\newblock \emph{ACM Computing Surveys (CSUR)}, 53(1): 1--37.

\bibitem[{Li et~al.(2023{\natexlab{a}})Li, Liang, Zhao, Zhang, Yang, and Xu}]{li2023stance}
Li, A.; Liang, B.; Zhao, J.; Zhang, B.; Yang, M.; and Xu, R. 2023{\natexlab{a}}.
\newblock Stance Detection on Social Media with Background Knowledge.
\newblock In \emph{Proceedings of the 2023 Conference on Empirical Methods in Natural Language Processing}, 15703--15717.

\bibitem[{Li et~al.(2023{\natexlab{b}})Li, Gao, Li, and Liao}]{li2023large}
Li, N.; Gao, C.; Li, Y.; and Liao, Q. 2023{\natexlab{b}}.
\newblock Large language model-empowered agents for simulating macroeconomic activities.
\newblock \emph{arXiv preprint arXiv:2310.10436}.

\bibitem[{Li et~al.(2021)Li, Sosea, Sawant, Nair, Inkpen, and Caragea}]{li2021p}
Li, Y.; Sosea, T.; Sawant, A.; Nair, A.~J.; Inkpen, D.; and Caragea, C. 2021.
\newblock P-stance: A large dataset for stance detection in political domain.
\newblock In \emph{Findings of the Association for Computational Linguistics: ACL-IJCNLP 2021}, 2355--2365.

\bibitem[{Liang et~al.(2022{\natexlab{a}})Liang, Chen, Gui, He, Yang, and Xu}]{liang2022zero}
Liang, B.; Chen, Z.; Gui, L.; He, Y.; Yang, M.; and Xu, R. 2022{\natexlab{a}}.
\newblock Zero-shot stance detection via contrastive learning.
\newblock In \emph{Proceedings of the ACM Web Conference 2022}, 2738--2747.

\bibitem[{Liang et~al.(2021)Liang, Fu, Gui, Yang, Du, He, and Xu}]{liang2021target}
Liang, B.; Fu, Y.; Gui, L.; Yang, M.; Du, J.; He, Y.; and Xu, R. 2021.
\newblock Target-adaptive graph for cross-target stance detection.
\newblock In \emph{Proceedings of the Web Conference 2021}, 3453--3464.

\bibitem[{Liang et~al.(2022{\natexlab{b}})Liang, Zhu, Li, Yang, Gui, He, and Xu}]{liang2022jointcl}
Liang, B.; Zhu, Q.; Li, X.; Yang, M.; Gui, L.; He, Y.; and Xu, R. 2022{\natexlab{b}}.
\newblock Jointcl: a joint contrastive learning framework for zero-shot stance detection.
\newblock In \emph{Proceedings of the 60th Annual Meeting of the Association for Computational Linguistics (Volume 1: Long Papers)}, volume~1, 81--91. Association for Computational Linguistics.

\bibitem[{Liang et~al.(2023)Liang, He, Jiao, Wang, Wang, Wang, Yang, Tu, and Shi}]{liang2023encouraging}
Liang, T.; He, Z.; Jiao, W.; Wang, X.; Wang, Y.; Wang, R.; Yang, Y.; Tu, Z.; and Shi, S. 2023.
\newblock Encouraging Divergent Thinking in Large Language Models through Multi-Agent Debate.
\newblock \emph{arXiv preprint arXiv:2305.19118}.

\bibitem[{Liu et~al.(2021)Liu, Lin, Tan, and Wang}]{liu2021enhancing}
Liu, R.; Lin, Z.; Tan, Y.; and Wang, W. 2021.
\newblock Enhancing zero-shot and few-shot stance detection with commonsense knowledge graph.
\newblock In \emph{Findings of the Association for Computational Linguistics: ACL-IJCNLP 2021}, 3152--3157.

\bibitem[{Lozhnikov, Derczynski, and Mazzara(2020)}]{lozhnikov2020stance}
Lozhnikov, N.; Derczynski, L.; and Mazzara, M. 2020.
\newblock Stance prediction for russian: data and analysis.
\newblock In \emph{Proceedings of 6th International Conference in Software Engineering for Defence Applications: SEDA 2018 6}, 176--186. Springer.

\bibitem[{Mohammad et~al.(2016)Mohammad, Kiritchenko, Sobhani, Zhu, and Cherry}]{mohammad2016semeval}
Mohammad, S.; Kiritchenko, S.; Sobhani, P.; Zhu, X.; and Cherry, C. 2016.
\newblock Semeval-2016 task 6: Detecting stance in tweets.
\newblock In \emph{Proceedings of the 10th international workshop on semantic evaluation (SemEval-2016)}, 31--41.

\bibitem[{Nair et~al.(2023)Nair, Schumacher, Tso, and Kannan}]{nair2023dera}
Nair, V.; Schumacher, E.; Tso, G.; and Kannan, A. 2023.
\newblock DERA: enhancing large language model completions with dialog-enabled resolving agents.
\newblock \emph{arXiv preprint arXiv:2303.17071}.

\bibitem[{OpenAI(2023)}]{openai2023gpt}
OpenAI, R. 2023.
\newblock GPT-4 technical report.
\newblock \emph{arXiv}, 2303--08774.

\bibitem[{Park et~al.(2023)Park, O'Brien, Cai, Morris, Liang, and Bernstein}]{park2023generative}
Park, J.~S.; O'Brien, J.~C.; Cai, C.~J.; Morris, M.~R.; Liang, P.; and Bernstein, M.~S. 2023.
\newblock Generative agents: Interactive simulacra of human behavior.
\newblock \emph{arXiv preprint arXiv:2304.03442}.

\bibitem[{Pontiki et~al.(2016)Pontiki, Galanis, Papageorgiou, Androutsopoulos, Manandhar, AL-Smadi, Al-Ayyoub, Zhao, Qin, De~Clercq et~al.}]{pontiki2016semeval}
Pontiki, M.; Galanis, D.; Papageorgiou, H.; Androutsopoulos, I.; Manandhar, S.; AL-Smadi, M.; Al-Ayyoub, M.; Zhao, Y.; Qin, B.; De~Clercq, O.; et~al. 2016.
\newblock Semeval-2016 task 5: Aspect based sentiment analysis.
\newblock In \emph{ProWorkshop on Semantic Evaluation (SemEval-2016)}, 19--30. Association for Computational Linguistics.

\bibitem[{Pontiki et~al.(2014)Pontiki, Galanis, Pavlopoulos, Papageorgiou, Androutsopoulos, and Manandhar}]{pontiki-etal-2014-semeval}
Pontiki, M.; Galanis, D.; Pavlopoulos, J.; Papageorgiou, H.; Androutsopoulos, I.; and Manandhar, S. 2014.
\newblock {S}em{E}val-2014 Task 4: Aspect Based Sentiment Analysis.
\newblock In \emph{Proceedings of the 8th International Workshop on Semantic Evaluation ({S}em{E}val 2014)}, 27--35. Dublin, Ireland: Association for Computational Linguistics.

\bibitem[{Qin et~al.(2023)Qin, Liang, Ye, Zhu, Yan, Lu, Lin, Cong, Tang, Qian et~al.}]{qin2023toolllm}
Qin, Y.; Liang, S.; Ye, Y.; Zhu, K.; Yan, L.; Lu, Y.; Lin, Y.; Cong, X.; Tang, X.; Qian, B.; et~al. 2023.
\newblock Toolllm: Facilitating large language models to master 16000+ real-world apis.
\newblock \emph{arXiv preprint arXiv:2307.16789}.

\bibitem[{Schick et~al.(2023)Schick, Dwivedi-Yu, Dess{\`\i}, Raileanu, Lomeli, Zettlemoyer, Cancedda, and Scialom}]{schick2023toolformer}
Schick, T.; Dwivedi-Yu, J.; Dess{\`\i}, R.; Raileanu, R.; Lomeli, M.; Zettlemoyer, L.; Cancedda, N.; and Scialom, T. 2023.
\newblock Toolformer: Language models can teach themselves to use tools.
\newblock \emph{arXiv preprint arXiv:2302.04761}.

\bibitem[{Shinn et~al.(2023)Shinn, Cassano, Labash, Gopinath, Narasimhan, and Yao}]{shinn2023reflexion}
Shinn, N.; Cassano, F.; Labash, B.; Gopinath, A.; Narasimhan, K.; and Yao, S. 2023.
\newblock Reflexion: Language Agents with Verbal Reinforcement Learning (arXiv: 2303.11366). arXiv.

\bibitem[{Tang et~al.(2020)Tang, Ji, Li, and Zhou}]{tang2020dependency}
Tang, H.; Ji, D.; Li, C.; and Zhou, Q. 2020.
\newblock Dependency graph enhanced dual-transformer structure for aspect-based sentiment classification.
\newblock In \emph{Proceedings of the 58th annual meeting of the association for computational linguistics}, 6578--6588.

\bibitem[{Touvron et~al.(2023)Touvron, Martin, Stone, Albert, Almahairi, Babaei, Bashlykov, Batra, Bhargava, Bhosale et~al.}]{touvron2023llama}
Touvron, H.; Martin, L.; Stone, K.; Albert, P.; Almahairi, A.; Babaei, Y.; Bashlykov, N.; Batra, S.; Bhargava, P.; Bhosale, S.; et~al. 2023.
\newblock Llama 2: Open foundation and fine-tuned chat models.
\newblock \emph{arXiv preprint arXiv:2307.09288}.

\bibitem[{Wang et~al.(2019)Wang, Shi, Kim, Oh, Yang, Zhang, and Yu}]{wang2019persuasion}
Wang, X.; Shi, W.; Kim, R.; Oh, Y.; Yang, S.; Zhang, J.; and Yu, Z. 2019.
\newblock Persuasion for good: Towards a personalized persuasive dialogue system for social good.
\newblock \emph{arXiv preprint arXiv:1906.06725}.

\bibitem[{Wang et~al.(2016)Wang, Huang, Zhu, and Zhao}]{wang2016attention}
Wang, Y.; Huang, M.; Zhu, X.; and Zhao, L. 2016.
\newblock Attention-based LSTM for aspect-level sentiment classification.
\newblock In \emph{Proceedings of the 2016 conference on empirical methods in natural language processing}, 606--615.

\bibitem[{Wei et~al.(2021)Wei, Bosma, Zhao, Guu, Yu, Lester, Du, Dai, and Le}]{wei2021finetuned}
Wei, J.; Bosma, M.; Zhao, V.~Y.; Guu, K.; Yu, A.~W.; Lester, B.; Du, N.; Dai, A.~M.; and Le, Q.~V. 2021.
\newblock Finetuned language models are zero-shot learners.
\newblock \emph{arXiv preprint arXiv:2109.01652}.

\bibitem[{Wei and Mao(2019)}]{wei2019modeling}
Wei, P.; and Mao, W. 2019.
\newblock Modeling transferable topics for cross-target stance detection.
\newblock In \emph{Proceedings of the 42nd International ACM SIGIR Conference on Research and Development in Information Retrieval}, 1173--1176.

\bibitem[{Wei, Mao, and Zeng(2018)}]{wei2018target}
Wei, P.; Mao, W.; and Zeng, D. 2018.
\newblock A target-guided neural memory model for stance detection in twitter.
\newblock In \emph{2018 International Joint Conference on Neural Networks (IJCNN)}, 1--8. IEEE.

\bibitem[{Xiang et~al.(2023)Xiang, Tao, Gu, Shu, Wang, Yang, and Hu}]{xiang2023language}
Xiang, J.; Tao, T.; Gu, Y.; Shu, T.; Wang, Z.; Yang, Z.; and Hu, Z. 2023.
\newblock Language Models Meet World Models: Embodied Experiences Enhance Language Models.
\newblock \emph{arXiv preprint arXiv:2305.10626}.

\bibitem[{Xu et~al.(2018)Xu, Paris, Nepal, and Sparks}]{xu2018cross}
Xu, C.; Paris, C.; Nepal, S.; and Sparks, R. 2018.
\newblock Cross-target stance classification with self-attention networks.
\newblock \emph{arXiv preprint arXiv:1805.06593}.

\bibitem[{Zeng et~al.(2022)Zeng, Liu, Du, Wang, Lai, Ding, Yang, Xu, Zheng, Xia et~al.}]{zeng2022glm}
Zeng, A.; Liu, X.; Du, Z.; Wang, Z.; Lai, H.; Ding, M.; Yang, Z.; Xu, Y.; Zheng, W.; Xia, X.; et~al. 2022.
\newblock Glm-130b: An open bilingual pre-trained model.
\newblock \emph{arXiv preprint arXiv:2210.02414}.

\bibitem[{Zhang, Ding, and Jing(2022)}]{zhang2022would}
Zhang, B.; Ding, D.; and Jing, L. 2022.
\newblock How would stance detection techniques evolve after the launch of chatgpt?
\newblock \emph{arXiv preprint arXiv:2212.14548}.

\bibitem[{Zhang et~al.(2023)Zhang, Fu, Ding, Huang, Li, and Jing}]{zhang2023investigating}
Zhang, B.; Fu, X.; Ding, D.; Huang, H.; Li, Y.; and Jing, L. 2023.
\newblock Investigating Chain-of-thought with ChatGPT for Stance Detection on Social Media.
\newblock \emph{arXiv preprint arXiv:2304.03087}.

\bibitem[{Zhang, Li, and Song(2019)}]{zhang2019aspect}
Zhang, C.; Li, Q.; and Song, D. 2019.
\newblock Aspect-based Sentiment Classification with Aspect-specific Graph Convolutional Networks.
\newblock In \emph{Proceedings of the 2019 Conference on Empirical Methods in Natural Language Processing and the 9th International Joint Conference on Natural Language Processing (EMNLP-IJCNLP)}, 4568--4578.

\bibitem[{Ziems et~al.(2023)Ziems, Held, Shaikh, Chen, Zhang, and Yang}]{ziems2023can}
Ziems, C.; Held, W.; Shaikh, O.; Chen, J.; Zhang, Z.; and Yang, D. 2023.
\newblock Can Large Language Models Transform Computational Social Science?
\newblock \emph{arXiv preprint arXiv:2305.03514}.

\end{thebibliography}

\clearpage

\section{Paper Checklist}

\begin{enumerate}

\item For most authors...
\begin{enumerate}
    \item  Would answering this research question advance science without violating social contracts, such as violating privacy norms, perpetuating unfair profiling, exacerbating the socio-economic divide, or implying disrespect to societies or cultures?
    \answerYes{Yes.}
  \item Do your main claims in the abstract and introduction accurately reflect the paper's contributions and scope?
    \answerYes{Yes.}
   \item Do you clarify how the proposed methodological approach is appropriate for the claims made? 
    \answerYes{Yes.}
   \item Do you clarify what are possible artifacts in the data used, given population-specific distributions?
    \answerNA{NA}
  \item Did you describe the limitations of your work?
    \answerYes{Yes, see the Conclusion and Future Work.}
  \item Did you discuss any potential negative societal impacts of your work?
    \answerNA{NA}
  \item Did you discuss any potential misuse of your work?
    \answerNA{NA}
    \item Did you describe steps taken to prevent or mitigate potential negative outcomes of the research, such as data and model documentation, data anonymization, responsible release, access control, and the reproducibility of findings?
    \answerYes{Yes, see the Experimental Setup.}
  \item Have you read the ethics review guidelines and ensured that your paper conforms to them?
    \answerYes{Yes.}
\end{enumerate}

\item Additionally, if your study involves hypotheses testing...
\begin{enumerate}
  \item Did you clearly state the assumptions underlying all theoretical results?
    \answerNA{NA}
  \item Have you provided justifications for all theoretical results?
    \answerNA{NA}
  \item Did you discuss competing hypotheses or theories that might challenge or complement your theoretical results?
    \answerNA{NA}
  \item Have you considered alternative mechanisms or explanations that might account for the same outcomes observed in your study?
    \answerNA{NA}
  \item Did you address potential biases or limitations in your theoretical framework?
    \answerNA{NA}
  \item Have you related your theoretical results to the existing literature in social science?
    \answerNA{NA}
  \item Did you discuss the implications of your theoretical results for policy, practice, or further research in the social science domain?
    \answerNA{NA}
\end{enumerate}

\item Additionally, if you are including theoretical proofs...
\begin{enumerate}
  \item Did you state the full set of assumptions of all theoretical results?
    \answerNA{NA}
	\item Did you include complete proofs of all theoretical results?
    \answerNA{NA}
\end{enumerate}

\item Additionally, if you ran machine learning experiments...
\begin{enumerate}
  \item Did you include the code, data, and instructions needed to reproduce the main experimental results (either in the supplemental material or as a URL)?
    \answerYes{Yes, see the Experimental Setup.}
  \item Did you specify all the training details (e.g., data splits, hyperparameters, how they were chosen)?
    \answerNA{NA}
     \item Did you report error bars (e.g., with respect to the random seed after running experiments multiple times)?
    \answerYes{Yes, we conduct multiple repeated experiments. In the main experimental results, we use a paired t-test when claiming that our method outperformed the best baseline.}
    \item Did you include the total amount of compute and the type of resources used (e.g., type of GPUs, internal cluster, or cloud provider)?
    \answerNA{NA}
     \item Do you justify how the proposed evaluation is sufficient and appropriate to the claims made? 
    \answerYes{Yes, see Experimental Setup and Experimental Results.}
     \item Do you discuss what is ``the cost`` of misclassification and fault (in)tolerance?
    \answerNA{NA}
  
\end{enumerate}

\item Additionally, if you are using existing assets (e.g., code, data, models) or curating/releasing new assets, \textbf{without compromising anonymity}...
\begin{enumerate}
  \item If your work uses existing assets, did you cite the creators?
    \answerYes{Yes, see the Experimental Setup and Experimental Results.}
  \item Did you mention the license of the assets?
    \answerYes{Yes.}
  \item Did you include any new assets in the supplemental material or as a URL?
    \answerYes{Yes, we provide the code for COLA.}
  \item Did you discuss whether and how consent was obtained from people whose data you're using/curating?
    \answerYes{No, because we only use open-sourced datasets.}
  \item Did you discuss whether the data you are using/curating contains personally identifiable information or offensive content?
    \answerYes{Yes, see the Datasets.}
\item If you are curating or releasing new datasets, did you discuss how you intend to make your datasets FAIR?
\answerNA{NA}
\item If you are curating or releasing new datasets, did you create a Datasheet for the Dataset? 
\answerNA{NA}
\end{enumerate}

\item Additionally, if you used crowdsourcing or conducted research with human subjects, \textbf{without compromising anonymity}...
\begin{enumerate}
  \item Did you include the full text of instructions given to participants and screenshots?
    \answerNA{NA}
  \item Did you describe any potential participant risks, with mentions of Institutional Review Board (IRB) approvals?
    \answerNA{NA}
  \item Did you include the estimated hourly wage paid to participants and the total amount spent on participant compensation?
    \answerNA{NA}
   \item Did you discuss how data is stored, shared, and deidentified?
   \answerNA{NA}
\end{enumerate}

\end{enumerate}

\end{document}